\documentclass{article}
\usepackage{spconf,amsmath,graphicx}

\usepackage{color}
\usepackage{xspace}
\usepackage{relsize}
\usepackage{caption}
\usepackage{subcaption}
\usepackage{multirow}
\usepackage{adjustbox}
\usepackage{wrapfig}

\newcommand\BLEU{\textsc{Bleu}\xspace}

\title{Two-Way Neural Machine Translation: A Proof of Concept for
Bidirectional Translation Modeling using a Two-Dimensional Grid}
\name{Parnia Bahar$^{1,2}$, Christopher Brix$^{1}$ and Hermann Ney$^{1,2}$}
\address{$^1$Human Language Technology and Pattern Recognition Group, Computer Science Department \\
RWTH Aachen University, 52074 Aachen, Germany, $^2$AppTek GmbH, 52062 Aachen, Germany \\
\texttt{\small \{bahar, ney\}@cs.rwth-aachen.de, christopher.brix@rwth-aachen.de}
    }

\begin{document}
%
\maketitle
\begin{abstract}
Neural translation models have proven to be effective in capturing sufficient information from a source sentence and generating a high-quality target sentence.
However, it is not easy to get the best effect for bidirectional translation, i.e., both source-to-target and target-to-source translation using a single model.
If we exclude some pioneering attempts, such as multilingual systems, all other bidirectional translation approaches are required to train two individual models.
This paper proposes to build a single end-to-end bidirectional translation model using a two-dimensional grid, where the left-to-right decoding generates source-to-target, and the bottom-to-up decoding creates target-to-source output.
Instead of training two models independently,
our approach encourages a single network to jointly learn to translate in both directions. Experiments on the WMT 2018 German$\leftrightarrow$English and Turkish$\leftrightarrow$English translation tasks show
that the proposed model is capable of generating a good translation quality and has sufficient potential to direct the research.
\end{abstract}

\section{Introduction \& Related Works} \label{sec:introduction}

Current state-of-the-art neural machine translation (NMT) systems are based on attention models \cite{bahdanau_15_attention,gehring_17_conv_seq2seq,vaswani_17_transformer} built on an encoder-decoder framework.
The encoder scans and transforms a source sequence into a sequence of vector representations, from which the decoder predicts a sequence of target words. 
Such systems are often referred to as unidirectional translation models as they translate only from one language (source) to another language (target).
Although such models have proven to be effective for a high-quality unidirectional translation, it is still challenging for an attention-based model to capture the intricate structural divergence between natural languages due to the non-isomorphism effect.
It is difficult to build a single system that translates reliably from and to two or even more languages.
In practice, usually, one model is trained for each direction, and each one might only capture partial aspects of the mapping between words \cite{cheng2016:ijcai2016:bidir_att}.
The two models seem to be complementary.
Therefore, combining the two models can hopefully improve translation quality in both directions.

A solution to use a single model to translate between multiple languages is a multilingual system. 
\cite{Dong2015:acl2015:multi} propose a one-to-many multilingual system that translates from one source language into multiple target languages by adding a separate attention mechanism and decoder for each target language.
\cite{Firat2016:naacl2016:multiway} apply a single shared attention mechanism but
multiple encoders and decoders for each source and target language, respectively.
Multi-task learning has also been proposed for many-to-many tasks \cite{Luong2016}, where the architecture is extended to have multiple encoders and decoders.
In the view of multilingual translation, each language in the source or target side is modeled by one separate encoder or decoder.
\cite{Johnson2017:tacl2017:multi_google} and \cite{Ha2016} have introduced a multilingual setting using a single attention encoder-decoder model. In such systems, no change in the architecture is needed, but they work with multilingual data to have both source-to-target and target-to-source translations involved. 
Such multilingual setups are beneficial for low-resource or even zero-shot scenarios.
These models require adding an artificial token to the input sequence to indicate the target language and depend on the network to identify the translation direction correctly.
Excluding the multilingual systems, \cite{cheng2016:ijcai2016:bidir_att} have proposed an attentional agreement by defining a new training objective that combines likelihoods in two directions with a word alignment agreement.

\begin{figure*}[t]
\centering
\includegraphics[width=0.70\textwidth]{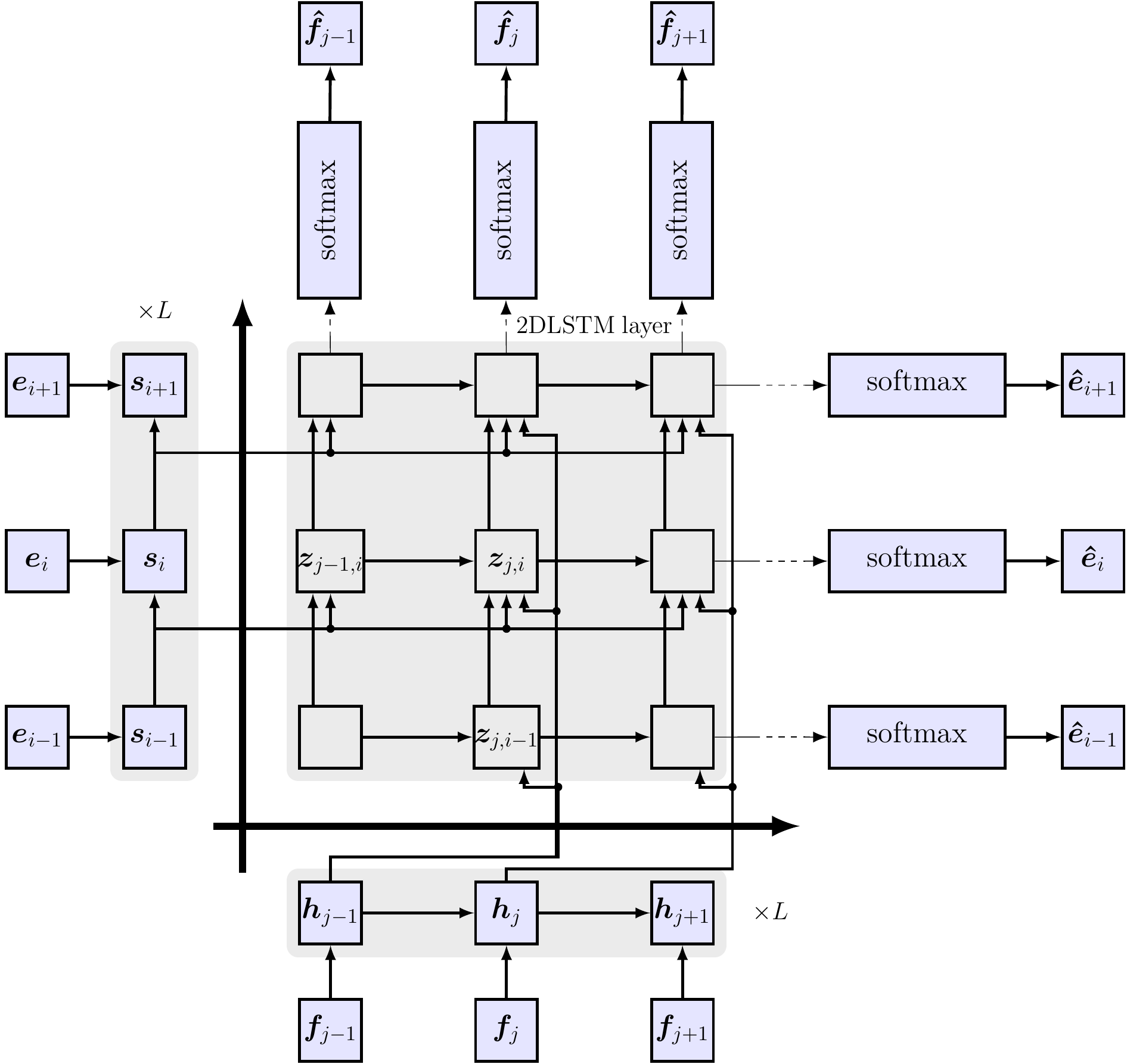}
\caption{Two-way bidirectional model using a 2DLSTM layer on top of $L$ layers of source and target encoders. Inspired by \cite{bahar2018:emnlp2018:nmt_2dseq}.}
\label{fig:2d_seq2seq}
\end{figure*}

Inspired by two-dimensional (2D) modelings \cite{bahar2018:emnlp2018:nmt_2dseq, bahar2019:icassp2019:asr_2dseq, ElbayadBV18}, we use a 2D grid 
to construct a single bidirectional translation model that opens up the interaction between source-to-target and target-to-source translation directions that are jointly trained on the same training data. 
In this architecture, we apply a 2D long short-term memory (LSTM) \cite{hochreiter_97_LSTM, Graves_2008_thesis} on top of the unidirectional source and target encoders to relate input and output representations in a 2D space. The horizontal axis of the 2DLSTM generates the source sentence, whereas the vertical dimension predicts the target sentence.
Our model is similar to an architecture used in machine translation described in \cite{bahar2018:emnlp2018:nmt_2dseq}, but it produces translations in both directions. We believe that the 2DLSTM can capture the correspondence between source and target words as well as the intricate structural divergence between two languages. 
Having a joint translation model working for both source-to-target and target-to-source benefits the research area in many aspects such as model latency and size, which recently got attention in machine translation \cite{frankleC19_lottery_ticket_nmt,brix_sparse_nmt}.
In the literature, we note that bidirectional NMT sometimes refers to left-to-right and right-to-left modeling, while here, it means source-to-target and target-to-source.

\section{Two-Way Bidirectional Translation Model}
\label{sec:2d seq2seq model}

Given a source sentence $f_1^{J}= f_1,\ldots, f_{J}$
that contains $J$ words and a target sentence $e_1^{I}= e_1,\ldots, e_{I}$ of $I$ words, 
the posterior probability of the target sequence $e_1^{I}$ is defined as $p(e_1^{I} |  f_1^{J})$.
As shown in Figure \ref{fig:2d_seq2seq}, we apply an encoder with $L$ layers ($L=4$) to scan the source sequence from left to right. The encoder layers can be composed of either unidirectional LSTM units or self-attention components \cite{vaswani_17_transformer}. 
In the latter case, to avoid the future context, we mask out the coming words and only attend on the histories up to time step $j$.
Similar to the source encoder, we use a stack of $L$ unidirectional (or masked) layers of target encoders on the target side.
The source and target sequence do not need to be of the same length. In Figure \ref{fig:2d_seq2seq}, we illustrate the same length for simplicity. 
The use of source and target encoders is optional in our architecture.
In other words, the source and target word embeddings can directly be used instead.
However, the initial experiments have shown that the additional encoding layers improve the performance. 
The encoder states are formulated as
\begin{align}
{h_j} &= \text{Enc}_{S}^{L} \circ \cdots \circ \text{Enc}_{S}^{1}(f_1^{j}), \nonumber \\
{s_i} &= \text{Enc}_{T}^{L} \circ \cdots \circ \text{Enc}_{T}^{1}(e_1^{i}),  \nonumber
\end{align}

where $h_j$ and $s_i$ are source and target encoder states computed by $\text{Enc}_{S}$ and $\text{Enc}_{T}$ functions respectively, which in our experiments are chosen to be masked multi-head self-attention layers.

\begin{figure*}[ht]
\centering
\begin{minipage}{0.3\textwidth}
  \centering
\includegraphics[width=0.8\textwidth]{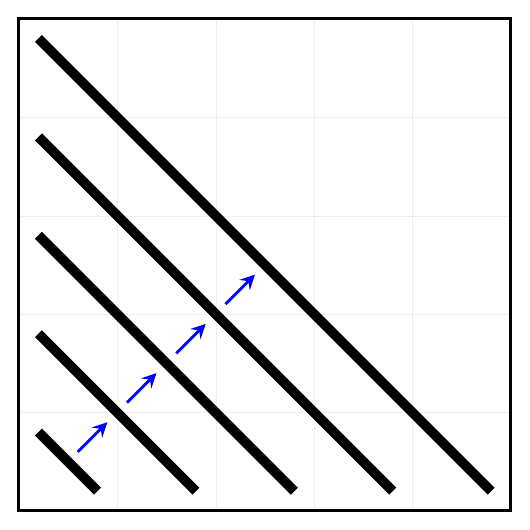}
\subcaption{diagonal-wise}\label{fig:diagonal}
\end{minipage}%
\begin{minipage}{0.3\textwidth}
  \centering
\includegraphics[width=0.8\textwidth]{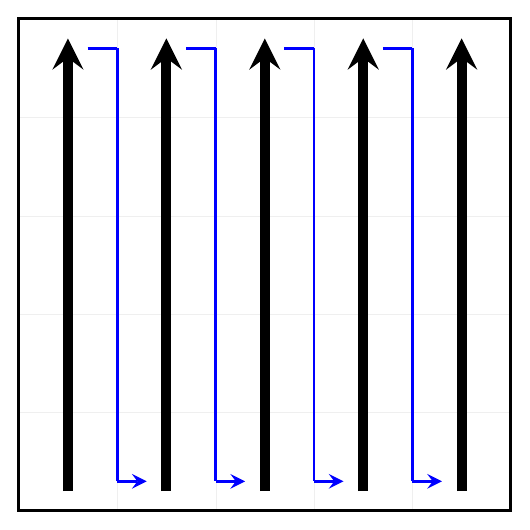}
\subcaption{column-wise}\label{fig:coumn}
\end{minipage}%
\begin{minipage}{0.3\textwidth}
  \centering
\includegraphics[width=0.8\textwidth]{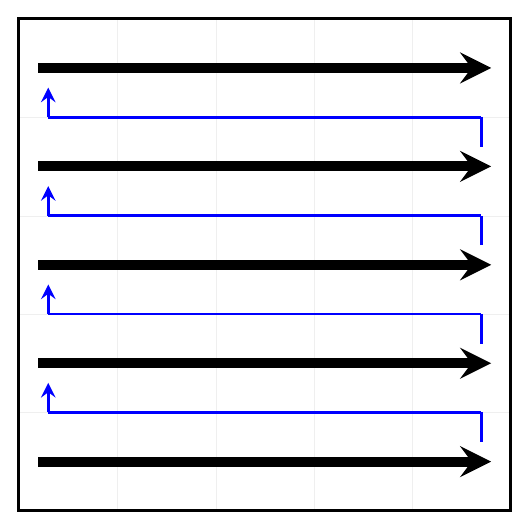}
\subcaption{row-wise}\label{fig:row}
\end{minipage}%
\caption{Order of processing. a) diagonal-wise used in training where all cells on the diagonal are computed in parallel. b) column-wise used in target-to-source decoding where columns are not computed at the same time. c) row-wise used in source-to-target decoding where rows are not computed at the same time. The black and blue arrows show the direction of a sequential process between cells and columns/rows,  respectively.} \label{fig:order_processing}
\end{figure*}

Similar to \cite{bahar2018:emnlp2018:nmt_2dseq, bahar2019:icassp2019:asr_2dseq}, we then equip the network with a 2D long short-term memory (2DLSTM) \cite{Graves_2008_thesis, Leifert_2016_stable_cells} layer to relate the source and target encoder states without any attention component in between.
A 2DLSTM unit has both horizontal and vertical recurrences, as illustrated in Figure \ref{fig:2d_seq2seq} that enable the cell to reconcile the context from both sides.
At time step $(j,i)$, the 2DLSTM receives a concatenation of the last source encoder state ${h}_{j-1}$, and the previous target encoder state ${s}_{i-1}$, as input.
Its recurrence relies on both the vertical ${z}_{j,i-1}$ and the horizontal hidden states ${z}_{j-1, i}$.
One dimension of the \mbox{2DLSTM} (horizontal-axis in the figure) sequentially reads the source encoder states from left to right and another (vertical axis) reads the target encoder states from bottom to top. 
The state of the 2DLSTM is given by

\begin{align}
{z}_{j,i} = \text{2DLSTM} \Big(\big[{h}_{j-1}; {s}_{i-1}\big], {z}_{j-1,i}, {z}_{j,i-1}\Big). \nonumber
\end{align}

The 2DLSTM state for a word at step $i$ only has a dependence on the preceding word sequence $e_1^{i-1}$.
Similarly, a state at step $j$ only depends on the preceding word sequence $f_1^{j-1}$.
At each decoder step, once the whole input sequence is processed from $1$ to $J$, we do max-pooling over all horizontal states to obtain the context vector. We have also tried average-pooling or taking the last horizontal state instead, but none performs better than max-pooling. 
In order to generate the next target word, ${e}_{i}$, a transformation followed by a softmax operation is applied. The same procedure is operated at the same time for the source word $f_j$.
Here, we pool over the vertical axis.
Therefore

\begin{align}
p(e_i| e_1^{i-1}, f_1^{J})  &= \text{softmax} \Big( \text{linear} \big(\text{pool} ({z}_{1,i}^{J,i}) \big) \Big)\big\rvert_{|V_e|}, \nonumber\\
p(f_j| f_1^{j-1}, e_1^{I})  &= \text{softmax} \Big( \text{linear} \big(\text{pool} ({z}_{j,1}^{j,I}) \big) \Big) \big\rvert_{|V_f|}, \nonumber
\end{align}

where $|V_f|$ and $|V_e|$ are the source and target vocabularies.
We note that we do not use any attention component in this model, and inherently we have no future context information (i.e., no bidirectional LSTM). The reason for this constraint is that including it on one language side breaks the network's ability to translate in this direction.
All the loss functions are differentiable with respect to model parameters. It is easy to extend the original training algorithm to implement joint training since the
two translation models in two directions share the same training data as well as the parameters. Our  training loss $\mathcal{L}$ is defined as

\begin{align}
\mathcal{L} = \log p(e_1^I| f_1^{J}) + \log p(f_1^{J}| e_1^{I}). \nonumber
\label{eq:multi:one2many}
\end{align}

\begin{table*}[th]
\begin{center}
\begin{adjustbox} {max width=1.0\textwidth}  
\begin{tabular}{l|p{3cm}|cccc}
Task & System  & test2015 & test2016  & test2017 & test2018\\ 
 \hline
\multirow{2}{*}{German$\to$English} 
&  transformer   & 32.4  & 37.5  & 33.4 & 40.4  \\
& bidir model  & 22.7 &  26.0 & 23.2 & 28.7  \\

 \hline
                
\multirow{2}{*}{English$\to$German} 
& transformer   & 28.5 &  33.4  & 27.2  & 40.2  \\
& bidir model  & 18.9 &  22.4 & 17.9 & 24.5 \\

\end{tabular}
\end{adjustbox}
\caption{Results measured in \BLEU${[\%]}$ score.}
\label{tab:results:de-en}
\end{center}
\end{table*}

\begin{table*}[th]
\begin{center}
\begin{adjustbox} {max width=1.0\textwidth}  
\begin{tabular}{l|p{3cm}|ccc}
Task & System & dev2016  & test2016 & test2017\\ 
 \hline
\multirow{3}{*}{Turkish$\to$English} 
& transformer  &   21.1 &	19.5  &  19.1   \\
& multilingual NMT  &  19.7  & 18.2	 &  18.1 \\
& bidir model  &   15.6   & 14.1 &  14.1  \\
 \hline
\multirow{3}{*}{English$\to$Turkish} 
& transformer  &  10.8  & 12.0 &  12.5  \\
& multilingual NMT  &  10.8 & 12.1 &  12.6  \\
& bidir model  &   \phantom{0}8.0  & \phantom{0}9.0 & \phantom{0}9.5 \\

\end{tabular}
\end{adjustbox}
\caption{Results measured in \BLEU${[\%]}$ score.}
\label{tab:results:tr-en}
\end{center}
\end{table*}

\subsection{Order of Processing}
During training, the entire source and target sentences are available.
This allows to compute all states of the \mbox{2DLSTM} before performing any pooling and prediction steps.
We process the 2D grid in a forward pass from the bottom left (step $(1, 1)$) to the top right corner (step $(J, I)$).
Afterward, vertical and horizontal slices can be used in parallel to compute the individual losses, which are then used for the backpropagation.
Importantly, the 2DLSTM state at timestep $(j, i)$ can only be processed after both predecessors at $(j-1, i)$ and $(j, i-1)$ are computed.
This constraint is fulfilled when processing the 2D grid diagonal-wise.
This processing scheme enables us to merge the kernel invocations for all positions on one diagonal, reducing the necessary computation time. 
As shown in Figure \ref{fig:diagonal},  all cells on the diagonal are computed in parallel. The blue arrows indicate a sequential process.
For a 2D grid with $J$ and $I$ words on the source and target sequence, respectively, there are $I \times J$ cells to process. In contrast, the number of diagonals is $I + J -1$, which in most cases should be considerably lower than the number of cells,  leading to faster training.
While the capacity of the GPU limits this parallelization, in practice, this reduces the training complexity to linear time \cite{Voigtlaender_2016_hwr}.
Similarly, the gradients are passed backward in the opposite direction, again using parallel computations for all cells on each diagonal.

In contrast, this optimization is not possible during decoding.
The translated sentence is not available, preventing the computation of multiple cells on the same diagonal at once.
Hence, we need to compute the states of the 2DLSTM row-wise for source-to-target, and column-wise for target-to-source translation (see Figure \ref{fig:order_processing}).
Here, for each column or row, the blue arrows imply that it depends on the one before that.

\section{Experiments} \label{sec:experiments}

\textbf{Dataset:} we carry out the experiments on two WMT translation tasks: German$\leftrightarrow$English and Turkish$\leftrightarrow$English, including 5.9M and 200K of sentence pairs, respectively.
After tokenization and true-casing using the \texttt{Moses} toolkit \cite{koehn_07_moses}, we  apply byte pair encoding (BPE) \cite{sennrich_16_bpe} with $50$k merge operations for the first and $20$k symbols for the second task. 

For German$\leftrightarrow$English and Turkish$\leftrightarrow$English, we use the \texttt{newstest2015} and the \texttt{newsdev2016} as the development set respectively the \texttt{newstest2016}, \texttt{2017} and \texttt{2018} as our test sets for German$\leftrightarrow$English while \texttt{newstest2016}, \texttt{2017} for the Turkish$\leftrightarrow$English. The models are evaluated using case-sensitive \BLEU \cite{papineni_02_bleu} computed by \texttt{mteval-v13a}.

\textbf{Model:} we train individual base transformers \cite{vaswani_17_transformer}
as our baseline systems for each direction. 
We use 6 layers in both the encoder and the decoder with internal dimension size of 512. 
We set the number of heads in the multi-head attention to 8. Layer normalization, dropout, and residual connections are applied. 
The models are trained end-to-end using the Adam optimizer with a learning rate of $0.0003$, and a dropout of $10\%$. 

Our bidirectional model has 4 layers of the masked multi-head self-attention encoder on both the source and target sides, where we masked the future tokens. It then behaves identically to unidirectional processing.
We use a learning rate of $0.0005$ and a dropout of $30\%$ on masked self-attention layers.
The 2DLSTM has 750 hidden units.
We also use an L2 norm of 0.05 for the 2DLSTM layer.
We employ a learning rate scheduling scheme, where we lower the learning rate with a decay factor of $0.9$ if the perplexity on the development set does not improve for several consecutive checkpoints. 
The maximum sequence length is set to 75 and 50 source tokens for the baseline and two-way models, respectively.
All batch sizes are specified to be as big as possible to fit in memory. 
A beam size of 12 is used in inference.
We use our in-house implementation of sequence to sequence modeling \texttt{RETURNN} \cite{zeyer_18_returnn}.
The code\footnote{https://github.com/rwth-i6/returnn} 
and the configurations of the setups are available\footnote{https://github.com/rwth-i6/returnn-experiments/}.

\section{Results} \label{sec:results}

The results can be seen in Table \ref{tab:results:de-en} for the German$\leftrightarrow$English and in Table \ref{tab:results:tr-en} for Turkish$\leftrightarrow$English tasks, respectively.
As shown in all directions, our bidirectional model underperforms the transformer baseline.
However, this difference is larger on the German$\leftrightarrow$English task compared to the low-resource scenario, where a bidirectional model works better.
On the English$\to$Turkish task, our two-way model is behind the baseline by almost 3\% \BLEU, which is the smallest gap.
These results indicate that building a single model that translates reliably for both directions is a difficult task.

We also set up a multilingual system for Turkish$\leftrightarrow$English based on the transformer model. Similar to \cite{Johnson2017:tacl2017:multi_google}, we concatenate the source and target side of training data with a special token for each direction. To translate from Turkish to English, we insert an \texttt{@en@} token, whereas for the reverse translation from English to Turkish, a \texttt{@tr@} token is added into the source data.
As listed in Table \ref{tab:results:tr-en}, the multilingual model slightly underperforms the transformer baseline on Tr$\to$En and outperforms it on En$\to$Tr. In comparison to our model, it results in better translation quality.

There are some potential reasons for such a huge gap between our model and the transformer model. 
Firstly, due to the inherent constraints of our model architecture, where we need to mask future tokens on both the source and target side, we have no access to the full context, and the model is unable to employ an encoder which can take into account unbounded bidirectional history.  It possibly is the main cause of the drop in performance.
Replacing the bidirectional encoder layers of the transformer architecture with unidirectional versions leads to a significant performance degradation\footnote{For example, an LSTM-based attention system with only unidirectional encoders gives 7.0\%  and 13.3\% in \BLEU on newsdev2016 for English$\to$Turkish and Turkish$\to$English respectively. While using a bidirectional encoder performs up to 12.0\% and 20.5\% in \BLEU on newsdev2016.}.
Secondly, the joint training for the bidirectional end-to-end NMT model requires a more sophisticated optimization compared to the independent training.
For joint training, where a new training objective combines likelihoods in two directions, we need to compromise between two tasks, and the parameters are updated jointly. In contrast to the separate training and updating of two independent sets of parameters, this might lead to a sub-optimal solution for the entire optimization problem, and such joined models are subject to more complex error propagation.

\section{Conclusion and Future Work} \label{sec:conclusion}

We proposed a two-way end-to-end bidirectional translation model, a single, yet joint (source, target) model based on a 2D grid. It permits source$\to$target and target$\to$source decoding along each axis, following joint training along both axes.
However, it is a work-in-progress paper, and more work might be needed to prove its effectiveness.
On a first attempt, the experimental results show that our architecture is able to generate reasonably good translations from source-to-target and target-to-source.

It has not yet reached parity on all tasks compared to separate models or a multilingual model in both directions using language tags; however, it offers a different and interesting modeling perspective. 
These are the first experiments using the 2DLSTM cell for the bidirectional translation modeling, and we expect better results with more tuning. More work needs to be done, and we intend to try the tasks with less reordering, such as translation between very related languages or paraphrasing.
Further exploration on a combination with non-autoregressive approaches is a correct research direction.
We also believe such an architecture motivates an alignment model where we can use bidirectional encoders on both source and target sides to align the words. The traditional alignment models, like GIZA++ \cite{giza} involve training models for both the directions and merging these bidirectional alignments afterward. We believe the two-way model with a combination of an attention mechanism is an appropriate candidate for such tasks where we are allowed to use bidirectional encoders. 

\section{Acknowledgements}
\label{sec:acknowledgements}
\begin{wrapfigure}{l}{0.10\textwidth}
\vspace{-8mm}
    \begin{center}
      \includegraphics[width=0.12\textwidth]{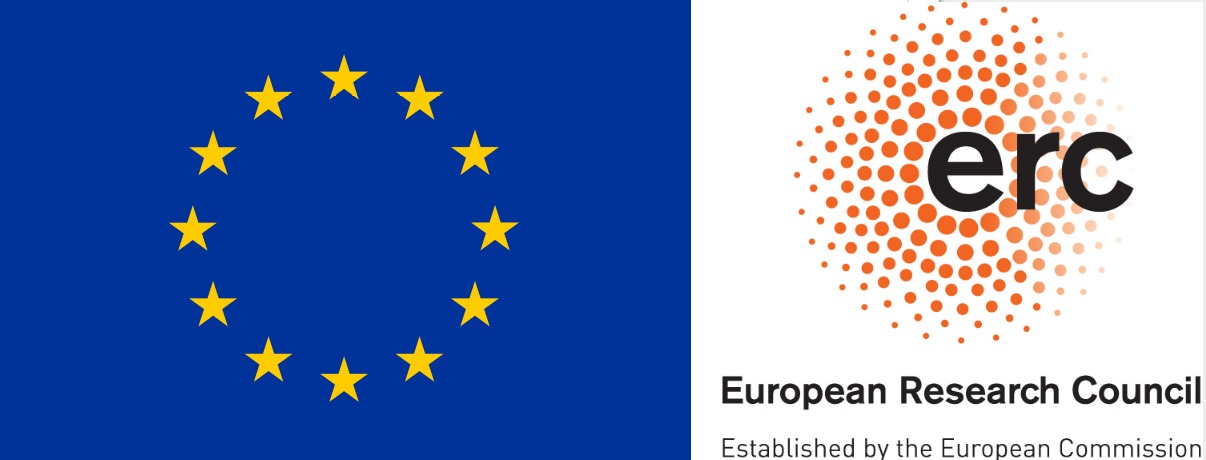} \\
      \vspace{2mm}
      \includegraphics[width=0.12\textwidth]{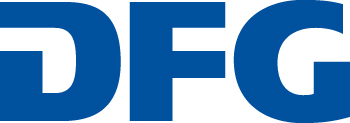}
    \end{center}
\vspace{-7mm}
\end{wrapfigure}
This work has received funding from the European Research Council (ERC) under the European Union's Horizon 2020 research and innovation programme (grant agreement No 694537, project "SEQCLAS"), the Deutsche Forschungsgemeinschaft (DFG; grant agreement NE 572/8-1, project "CoreTec"). The work reflects only the authors' views and none of the funding parties is responsible for any use that may be made of the information it contains.



\bibliographystyle{IEEEbib}



\let\OLDthebibliography\thebibliography
\renewcommand\thebibliography[1]{
  \OLDthebibliography{#1}
  \setlength{\parskip}{0pt}
  \setlength{\itemsep}{0pt plus 0.07ex}
}

\bibliography{refs}

\end{document}